\documentclass[runningheads]{llncs}

\usepackage{pdfpages}
\usepackage{graphicx}
\usepackage{mathtools}
\usepackage{soul}
\usepackage{xcolor}
\usepackage{booktabs}
\usepackage{amsfonts}
\usepackage{array,multirow}
\usepackage{hyperref}
\hypersetup{
    colorlinks=true,
    linkcolor=black,
    filecolor=black,      
    urlcolor=blue,
    citecolor=black,
}

\newcommand{\STAB}[1]{\begin{tabular}{@{}c@{}}#1\end{tabular}}

\begin{document}
\title{Variational Shape Completion for Virtual Planning of Jaw Reconstructive Surgery
}
\titlerunning{Variational Shape Completion}
%
\author{Amir H. Abdi \and
Mehran Pesteie \and
Eitan Prisman \and
Purang Abolmaesumi \and
Sidney Fels}

%
\authorrunning{Abdi and Pesteie et al.}
%
\institute{
University of British Columbia, Vancouver, Canada
\email{\{amirabdi,mehranp,purang,ssfels\}@ece.ubc.ca, eitan.prisman@ubc.ca}
}
\maketitle              
\begin{abstract}
The premorbid geometry of the mandible is of significant relevance in jaw reconstructive surgeries and occasionally unknown to the surgical team. In this paper, an optimization framework is introduced to train deep models for completion (reconstruction) of the missing segments of the bone based on the remaining healthy structure. 
To leverage the contextual information of the surroundings of the dissected region, the voxel-weighted Dice loss is introduced. 
To address the non-deterministic nature of the shape completion problem, we leverage a weighted multi-target probabilistic solution which is an extension to the conditional variational autoencoder (CVAE). 
This approach considers multiple targets as acceptable reconstructions, each weighted according to their conformity with the original shape. We quantify the performance gain of the proposed method against similar algorithms, including CVAE, where we report statistically significant improvements in both deterministic and probabilistic paradigms. 
The probabilistic model is also evaluated on its ability to generate anatomically relevant variations for the missing bone. 
As a unique aspect of this work, the model is tested on real surgical cases where the clinical relevancy of its reconstructions and their compliance with surgeon's virtual plan are demonstrated as necessary steps towards clinical adoption.

\end{abstract}
\section{Introduction}

Head and neck cancer comprises of a set of malignant tumors in the upper respiratory tract  which constitutes 3-4\% of cancer cases in North America~\cite{cancerUS2017}.
Surgery is the first line of treatment for the majority of these cases.
Despite the advances in tools and techniques,
mandibular reconstruction after 
segmental mandibulectomy is still a challenging procedure.
Moreover, due to the mandible's vital role in mastication, speech, and swallowing, the functional and aesthetic requirements of mandibular reconstructions are quite high.

The vascularized free fibula flap is the most utilized technique for mandibular reconstruction~\cite{FFF} where the linear shape of the fibula bone is dissected, contoured, and modelled to complete the curved geometry of the ablated bone.
In the past decade,  three-dimensional~(3D) virtual surgical planning (VSP) has gained traction.
In the VSP-enhanced free fibula flap, the modelling and shaping of the fibula are virtually planned based on the pre-operative records.

One of the challenges in mandibular reconstruction is the unknown pre-incident shape of the missing (distorted) bony elements.
In cases where the shape of the target bony element is unknown, 
three strategies are taken to anticipate the original anatomy~\cite{MandibleReconst2019}.
In rare cases where previous medical records of the craniofacial skeleton is available, patient's own bone morphology serves as the reference.
For unilateral defects, the uninvolved contralateral side is mirrored and manually positioned to create an estimated reconstruction reference.
However, this approach becomes less reliable for defects with anteromedial extensions and is inapplicable in midline crossing lesions.
Standard anatomic templates are seldom utilized to fill the missing bony elements after subjet-specific adjustments.

\subsubsection{Related Work}
\label{sec:orgc0d403a}
Shape completion is an ill-posed inverse problem in computer vision and graphics.
Data-driven models, including deep neural networks,  
are more intriguing as they directly learn the completion under supervision particularly because of the non-deterministic ground-truth and multiple acceptable solutions for the reconstruction problem.
In the realm of 3D generative deep models, 
applicability of adversarial training (GAN) in 3D shape and point cloud syntheses are investigated~\cite{achlioptas2017representation,wu2016learning}.
Varitional autoencoders (VAE) have also been able to learn semantically meaningful latent spaces for realistic completions~\cite{nash2017shape}.
The closest research to our work is a recent study where they empirically proved the feasibility of generating mandible  shapes from a predefined set of landmarks~\cite{Abdi2019}.

\subsubsection{Contributions}
We introduce optimization approaches to train deep convolutional models to reconstruct the missing bone segment from the remaining healthy mandible. 
Our contributions are three-fold. 
First, we design a  framework to randomly generate samples for training of shape completion models in deterministic and probabilistic paradigms.
We introduce the Voxel-weighted Dice loss that prioritizes the target (removed) region over the rest of the geometry while ensuring consistency in reconstruction by leveraging the contextual information from the rest of the shape.
Our main contribution is the Target-weighted variational  objective function which addresses the one-to-many reality of shape reconstruction.
We report the quantified gain in  performance and demonstrate the model's ability to predict variations of the missing bone.
Moreover, as a unique aspect of this work, qualitative results on real surgical cases are provided as a notion of  clinical relevancy  and compliance with surgeon's virtual plans.

\section{Method}


\subsection{Network Architecture}

\begin{figure}
\includegraphics[trim={0 0 0 0},clip ,width=1.0\textwidth]{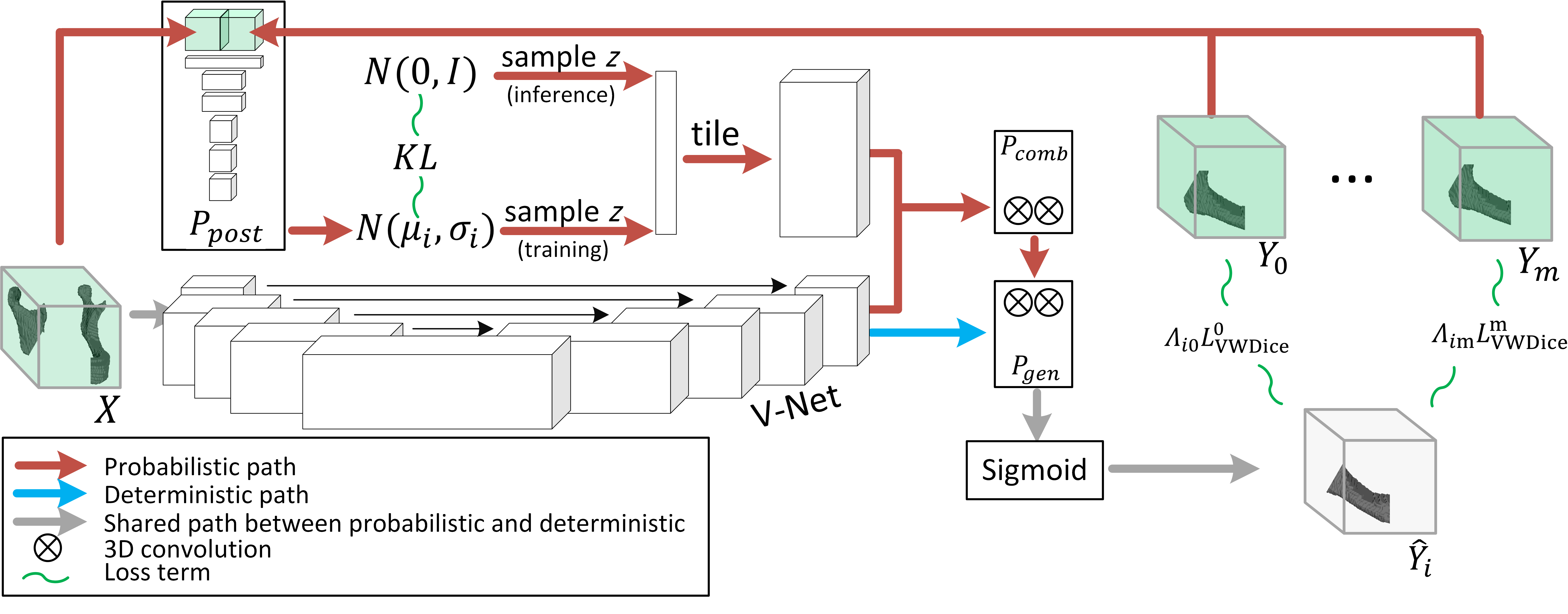}
\caption{Architecture of the deterministic and weighted multi-target probabilistic learning framework for anatomy completion with 3D convolutional models. 
} \label{fig:architecture}
\end{figure}

The input ($X$) to the deep model is a 3D binary voxel-grid of size $c^3$ containing a mandible with a missing (dissected) contiguous segment.
The output of the model is a 3D probability map ($\hat{Y}$) of  size $c^3$ where the predicted volume is the union of all voxels with $Y_{ijk} > 0.5$.
Architecture of the learning system is summarized in Fig.\ref{fig:architecture}.
We investigate two reconstruction approaches, deterministic and probabilistic.
In heart of both lies a V-Net~\cite{vnet}
consisting of four down-transition and four up-transitions with skip connections to forward feature maps from the encoding to the decoding stream.  Each down-transition consists of strided 3D convolutions and batch-normalization.
Similarly, the up-transitions contain transposed strided 3D convolution (deconvolution) and batch-normalization.
ELU activation layers are used throughout the V-Net.
Activations of the main stream of the last two down-transitions and the first two up-transitions are dropped out with a probability of 0.5.
The final up-transition generates four 3D feature maps of size  $c^3$, same size as the input and the target occupancy maps. 
More details of the architecture are available in the supplementary material.


In deterministic shape reconstruction, the final four feature maps are convolved with $P_{gen}$ kernels of size 2 followed by a sigmoid function.
In the probabilistic approach, the four feature maps of the V-Net are first concatenated with the tiled latent values (see Section~\ref{sec:loss-functions}) and convolved with 8 kernels in two consecutive layers of $P_{comb}$ prior to $P_{gen}$~(Fig.~\ref{fig:architecture}).

\subsection{Loss Functions}
\label{sec:loss-functions}

During the mini-batch gradient decent optimization, each training sample is randomly dissected by a cuboid ($B$) with an arbitrary size and orientation, which forms a binary occupancy map inside a 3D grid of size $c^3$. 
The cuboid cuts the mandible to create input and target shapes as follows,
\begin{equation}
X = S \circ B'   \quad \text{ and } \quad   Y = S \circ B,
\end{equation}
where $P \circ Q$ denotes the Hadamard product of the tensors $P$ and $Q$, and  $B'$ is the complement of the binary cuboid.
$X$ and $Y$ are the input and target  (Fig.~\ref{fig:variations}).


\subsubsection{Voxel-weighted Dice}

This objective function enables the model to leverage the contextual information from the surroundings of the dissected region.
Hence, voxels of the predicted 3D probability map are weighted adversely by their distance from the removed bony segment.
To do so, a 3D normal distribution, 
$\mathcal{N}_w(\bar{Y}, \sigma_w^2 I)$,
with a diagonal covariance matrix is instantiated at the center of the target segment ($\bar{Y}$). The 3D weight matrix $W$ of size $c^3$ is then defined as
\begin{equation}
    W_{ijk}= 
    \begin{cases}
    1 &\quad B_{ijk}=1 \\ 
    (2\pi\sigma_{w})^{\frac{3}{2}} \mathcal{N}_w(i,j,k)  
    &\quad B_{ijk}=0 \\
    \end{cases}.
\label{eq:w}
\end{equation}
Eq.~\ref{eq:w} prioritizes the voxels inside the volume of the dissection cuboid~($B$) and ignores distant voxels.
Here, $\sigma_w$ is set to $c/3$ to encourage smoother contours.

Based on the weight matrix, $W$, the proposed Voxel-weighted Dice loss is
\begin{equation}
    \mathcal{L}_{\text{VWDice}}(X,Y,\hat{Y},W) = 
    1 - \frac{2 * \sum\limits_{ijk} 
    \big[\hat{Y} \circ (Y+X) \circ W \big]_{ijk}}
    {\sum\limits_{ijk} 
    \big[(\hat{Y} + Y+X)  \circ W \big]_{ijk}} .
\label{eq-wDice}
\end{equation}
The proposed Voxel-weighted Dice prioritizes the dissected region for shape completion and maintains consistency between the synthesized structure~($\hat{Y}$) and the remaining anatomy~($X$).

\subsubsection{Target-weighted Loss}

Shape completion in the anatomical domain is inherently a non-deterministic process,
\emph{i.e.}, there is no single ground-truth that completes the dissected input anatomy. 
Therefore, our objective is to learn the variations of the removed bony segment from a dataset of mandibles and generate multiple solutions to reconstruct a given dissection.
In the proposed probabilistic learning paradigm, 
training samples are randomly dissected, on-the-fly, and the removed segment is considered the best known reconstruction target, referred to as $Y_0$, hereafter.
In the proposed multi-target approach, $m$ random training samples  are  selected and dissected at the same location using the same cuboid and registered on $Y_0$ to create the set of possible reconstructions,
$\mathcal{Y} = \big\{ Y_{0}, Y_{1}, ..., Y_{m} \big\} $.
Among $\mathcal{Y}$, only $Y_{0}$ perfectly completes $X$ while others do not necessarily match the input.
Therefore, the metric $\Lambda_{ij}(\mathcal{Y})$ is defined as the degree of geometrical conformity between members $i$ and $j$ of the set $\mathcal{Y}$.


All target nominees ($\mathcal{Y}$) are then independently concatenated with $X$ and mapped to the latent position $\mu_i$, with uncertainty $\sigma_i$, via the posterior encoder network ($P_{post}$).
The resultant mappings are considered as parameters of an independent multivariate normal distribution~($\mathcal{N}(\mu_i, \sigma_i)$).
A sample from this distribution is tiled and concatenated with the feature maps of the V-Net.
The resultant 4D tensor is processed by $P_{comb}$ and $P_{gen}$ to predict the target:
\begin{equation}
    \hat{Y_i} = P_{gen}(P_{comb} (V_{Net}(X), \mu_i)), \quad 
    P_{post}(.|X, Y_i) = \mathcal{N}(\mu_i, \sigma_i) .
\end{equation}
This process is repeated for all  $m+1$ nominee targets in  the $\mathcal{Y}$ set of sample $S$.

Each predicted shape ($\hat{Y}_i$) is compared with its corresponding target ($Y_i$) using the Voxel-weighted Dice.
Since target nominees, other than $Y_{0}$, involve deviations from the input morphology ($X$),
each target's conformity ($\Lambda_{i0}$) with the best known solution ($Y_0$)
is taken into account in the proposed Target-weighted (TW) loss function.
As a result, the final objective function is the weighted average of all loss functions with respect to the conformities of their targets.
The proposed objective is formulated as follows:
\begin{equation}
\label{eq:vwDice-tw}
  \mathcal{L}_{\text{VWDice-TW}} =  
\alpha
  \sum_{i=0}^{m}
  \Lambda_{0i}(\mathcal{Y})~ \mathcal{L}^i_{\text{VWDice}}(X,Y_i,\hat{Y_i},W_i) +
  \gamma KL(P_{post}||\mathcal{N}(0,I)) ,
\end{equation}
where, $\alpha$ is a normalizing parameter set as the sum of all $m$ conformity values, and $\gamma$ is a weighting constant. 
As shown in Eq.~\ref{eq:vwDice-tw}, during optimization, the Kullback\textendash Leibler (KL) divergence of the posterior latent distribution with a fixed normal distribution, from which we sample during inference, is minimized.

The $\mathcal{L}_\text{VWDice-TW}$ loss function considers all targets as partially acceptable solutions for the probabilistic completion.
In our experiments, $\Lambda(.)$ was set as the Dice coefficient between the shapes, where clearly $\Lambda_{00}(\mathcal{Y}) = Dice(Y_0, Y_0) = 1$.


\section{Experiments}

\subsection{Data and Training}
A total of 117  surface meshes of healthy mandibles were collected from three publicly available sources~\cite{miccai-challenge,wallner_10samples,tcia-tcga}. 
Some mandibles  from the archives of an anonymized center were also included through  data sharing agreements. 

The 3D meshes were rigidly registered based on their point clouds using the group-wise student's-t mixture model algorithm with 50 mixture components~\cite{Ravikumar2016}.
Using ray testing, each surface  mesh was voxelized into an isotropic binary occupancy voxel map with 1~mm increments to mimic mandibles segmented from CT scans with isotropic voxels.
The occupancy maps of all mandibles were symmetrically zero padded to create voxel-grid cubes of size $141^3$, \emph{i.e.} size of the largest sample, which also matches the maximum  facial width reported in the comprehensive dataset of the FaceBase project~\cite{Brinkley2016}. 
The dataset was randomly partitioned into test (15\%), training (70\%), and validation (15\%) sets.

During the mini-batch gradient decent optimization, each sample was randomly rotated, translated (shifted), and mirrored across the sagittal plane. 
Adam optimizer was used with default momentum parameters along with $\ell_2$ regularization of $1e-5$.
The learning rate was initialized at $1e-2$ and exponentially decayed with a rate of 0.98 at each epoch until convergence.
Size of the latent space was set to 8.
Same random seeds were used for all the experiments

The data processing pipeline and the  models were implemented using the PyTorch deep learning platform and made publicly available: \href{http://github.com/amir-abdi/prob-shape-completion}{github.com/amir-abdi/prob-shape-completion}.
For the experiments to be reproducable, 
the voxelized version of the data  accompanies the code, according to each dataset's respective license and data sharing agreements.

\subsection{Evaluation and Results}
\label{sec:results}

\begin{table}[tb]
\centering
\setlength{\tabcolsep}{3pt}
\caption{Quantitative comparison of the proposed methods against other baselines. }
\begin{tabular}{l|lcccc}  
\cline{2-6}
 & 
 \multicolumn{1}{c}{Method} & 
 \multicolumn{1}{c}{DSC\%} & 
 \multicolumn{1}{c}{Comp} & 
 \multicolumn{1}{c}{Acc} & 
 \multicolumn{1}{c}{HD95} \\ \hline
\multicolumn{1}{|c| }{\multirow{3}{*}{\STAB{\rotatebox[origin=c]{90}{ Determ.}}}}
& $\mathcal{L}_{Dice}(X+Y,\hat{Y})$ &   $0.4 $    & N/A & N/A & N/A \\
\multicolumn{1}{|c| }{}
& $\mathcal{L}_{Dice}(Y,\hat{Y})$ & 
                    $84.9 \pm 0.2 $    & 
                    $0.85 \pm 0.13 $ & 
                    $0.61 \pm 0.13$   & 
                    $2.95 \pm 1.43$   \\
\multicolumn{1}{|c| }{}
& $\mathcal{L}_{\text{VWDice}}(X,Y,\hat{Y})$ $_\text{(ours)}$ & 
                    $\mathbf{88.3 \pm 0.2} $    & 
                    $\mathbf{0.65 \pm 0.10}$ & 
                    $\mathbf{0.44 \pm 0.10}$   & 
                    $\mathbf{2.64 \pm 1.83}$  \\
\midrule
\midrule
\multicolumn{1}{|c| }{\multirow{2}{*}{\STAB{\rotatebox[origin=c]{90}{Prob.}}}}
& CVAE$_\text{basic}$ \cite{cvae} & 
                    $79.8 \pm 0.6 $    & 
                    $1.20 \pm 0.34 $ & 
                    $0.87 \pm 0.18$   & 
                    $3.98 \pm 3.16$   \\
\multicolumn{1}{|c| }{}
& CVAE$_{\text{VWDice-TW}}$  $_\text{(ours)}$& 
                    $\mathbf{80.8 \pm 0.5} $    & 
                    $\mathbf{1.11 \pm 0.31} $ & 
                    $\mathbf{0.83 \pm 0.14}$   & 
                    $\mathbf{3.74 \pm 2.44}$   \\
\bottomrule
\end{tabular}
\label{table:results}
\end{table}

Shape completion is a non-deterministic process where no single answer is the ground-truth.
However, to quantify the performance of the proposed learning methods, a dentist manually removed bone segments from the healthy mandibles of the test set and created nearly 100 test cases.
The manually removed bone segments were considered as fair targets for evaluation
and compared with the predicted reconstructions.
We assessed the models based on Dice coefficient (DSC; $2 \times \text{intersection} / \text{sum of volumes}$), completeness (Comp; average distance from target surface to predicted surface), accuracy (Acc; average distance from predicted surface to target surface), 
and Hausdorff distance at the 95th percentile~(HD95). 
Except for the DSC, lower values of the metrics are preferred.
For a fair comparison of non-deterministic CVAE models, and with only a single target reconstruction available, latent values during the quantitative analysis were set to the mean of the fixed distribution $\mathcal{N}(0, I)$.

\begin{figure}[tbp]
\centering
\includegraphics[trim={0 0 0 0},clip ,width=0.8\textwidth]{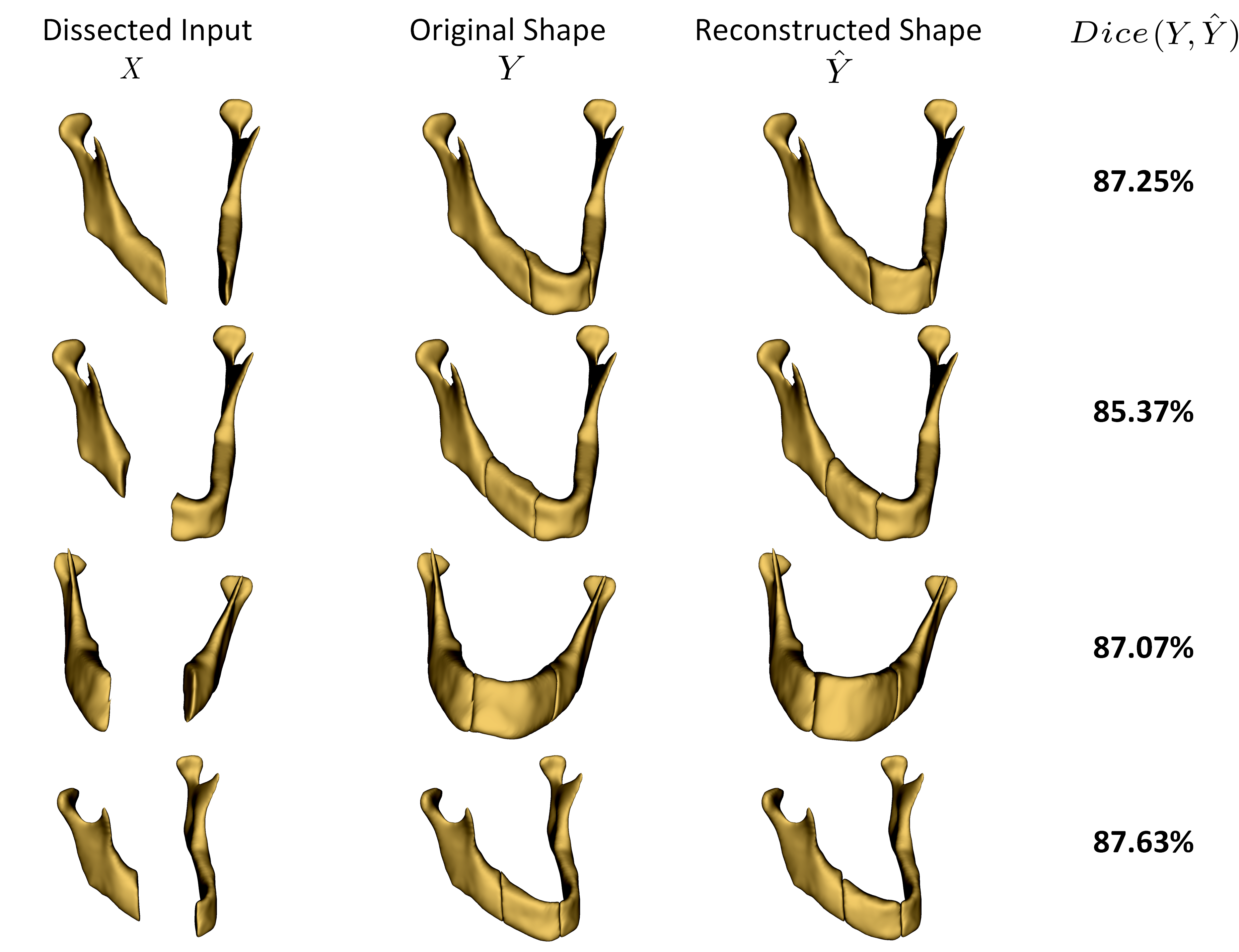}
\caption{Reconstructed samples of the test set along with their Dice metric when compared with the original anatomy.} 
\label{fig:samples}
\end{figure}

The quantitative results are reported in Table~\ref{table:results} and a set of reconstructed samples from the test set are visualized in Fig.~\ref{fig:samples}.
As reported in the top row of Table~\ref{table:results}, a vanilla Dice loss with the entire shape as its target~(\emph{i.e.} $L_{Dice}(X+Y,\hat{Y})$) 
equally treats all the voxels and ignores the dissected region. This model acts like an autoencoder which only regenerates the input ($X$).
On the contrary, focusing only on the dissected target bone~(\emph{i.e.} $L_{Dice}(Y,\hat{Y})$) does not penalize the  discrepancies in the margins of the reconstruction. Therefore, the predicted shape shows inconsistencies and discontinuities with respect to the surrounding anatomy.
The performance gain   achieved with the proposed objective function (Eq.~\ref{eq-wDice}) was assessed to be statistically highly significant~($p < 0.001$).

To demonstrate the effectiveness of the proposed contributions, 
two CVAE models were trained: 
one without any enhancements ($\text{CVAE}_\text{basic}$), 
and one with the objectives function described in Eq.~\ref{eq-wDice} and \ref{eq:vwDice-tw} ($\text{CVAE}_\text{VWDice-TW}$).
Network architectures and hyper-parameters were kept the same  across experiments for results to be comparable.
The differences between the Dice metrics of the two models was observed to be statistically highly significant ($p < 0.001$).

\begin{figure}[tbp]
\centering
\includegraphics[trim={0 15 0 0},clip ,width=0.99\textwidth]{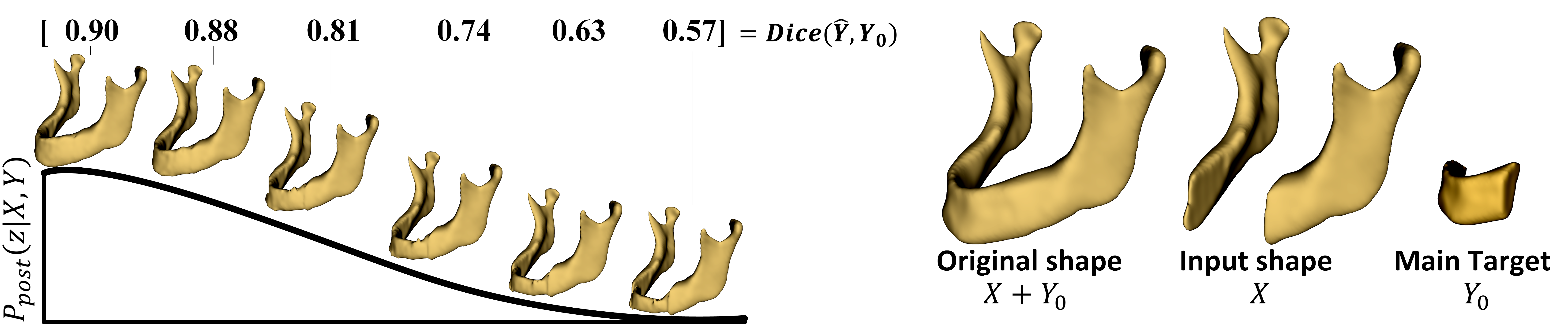}
\caption{Negative correlation between the deviation  of the latent values from the mode~($\hat{P_{post}}$) and the conformity of the predicted shapes with the main target ($Y_0$).} 
\label{fig:variations}
\end{figure}

Thanks to the proposed Target-weighted variational objective, a strong negative correlation was observed between the deviation of a latent vector from its distribution's mode ($|z - \hat{P}_{post} |$) and the similarity of its corresponding predicted shape with the main target ($Y_0$). This phenomenon is demonstrated in Fig.~\ref{fig:variations}.
The same was not true for the CVAE$_\text{basic}$  model.


\subsubsection{Surgical Cases}
\begin{figure}[tbp]
\centering
\includegraphics[trim={35 100 35 150},clip ,width=0.8\textwidth]{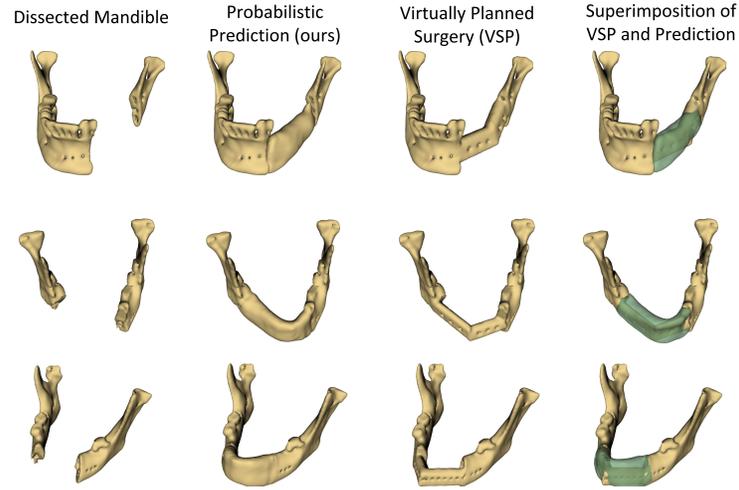}
\caption{Comparison of model predictions (green) with virtual surgical plans (VSPs).} 
\label{fig:surgery}
\end{figure}

We evaluated the performance of the variational model on real surgical  cases who were treated with the virtually planned free fibula flap technique. 
Here, the affected mandibular bones were already removed by the clinicians.
The dissected mandibles were reconstructed using the trained variational model.
An expert assessed the reconstructions against surgeon's virtual plans of the free fibula flap surgery and found them clinically acceptable.
Figure~\ref{fig:surgery} visualizes some of these surgical cases by superimposing the predicted mandibular bone (green)  on the fibular segments.


\section{Discussion and Conclusions}
In this paper, we introduced optimization approaches for training of deep  variational models  for anatomical 3D shape completion.
The proposed Voxel-weighted objective improves the accuracy of reconstructions compared to similar approaches and  guarantees smoothness between the predicted bony segment and the remaining contours of the mandible.
The proposed variational method takes into account the many acceptable solutions for the shape completion problem and is able to generate realistic variations for the missing segment.

Among the limitations of our study is the inconsistency in the presence of teeth across the samples. While there were toothless mandibles in the dataset, the majority of samples had all or parts of their dentition. 
However, except for their correlation with the bone-loss, teeth  have little to no role to play in determining the missing anatomy of mandible. 
Therefore, the presence of dentition in the target dissections adversely affected the performance. 
To mitigate this issue teeth should be excluded from the data, either manually or automatically.

The proposed probabilistic approach is among the first works in deep anatomical shape reconstruction. It can be applied to other anatomies as well as general computer graphics. 
Comparison with real surgical cases is demonstrated here as a step for clinical adoption.
Our method requires no post processing, except in converting voxel-grids to and from surface meshes.
Therefore, our next logical step is to switch from voxel-grids to graph representations to speedup the processing pipeline for better clinical acceptance.


\bibliographystyle{splncs04}
\bibliography{mybib}

\newpage
\appendix

\includepdf[pages=-,scale=0.95,offset=0 0]{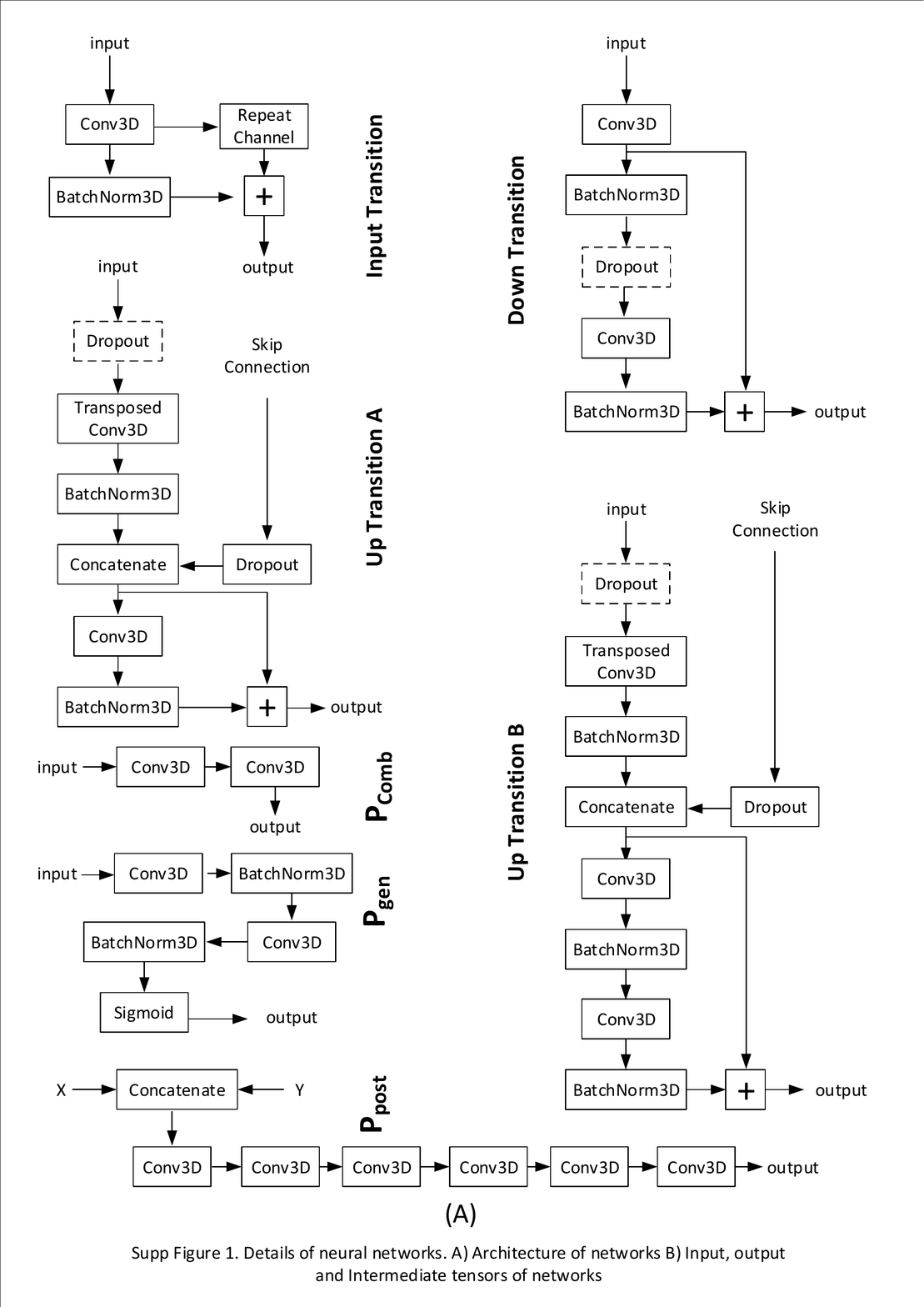}

\end{document}